%% file: main.tex
\documentclass[conference]{IEEEtran}
\IEEEoverridecommandlockouts

\input{include/includes}

\input{include/setup}
\input{include/acronyms}

\usepackage{kamomargins}
\kamoInit{}
\kamoSetMargins{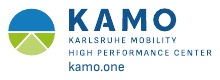}
\kamoSetIEEEFoot{2025}
\kamoSetIEEEHead{Lüttner, F., Neis, N., Stadler, D., Moss, R., Fehling-Kaschek, M., Pfriem, M., ... \& Ziehn, J. (2025). From Real-World Traffic Data to Relevant Critical Scenarios. In International Automated Vehicle Validation Conference (IAVVC) 2025}
{10.1109/IAVVC61942.2025.11219576}

\newcommand\TTCE{\mathit{TTCE}}

\newcommand\DCE{\mathit{DCE}}
\newcommand\THW{\mathit{THW}}
\newcommand\vlim{v\subs{lim}}

\begin{document}

\title{From Real-World Traffic Data\\to Relevant Critical Scenarios
\thanks{This work was funded by the German Federal Ministry for Economic Affairs and Climate Action (BMWK) within the program ``New Vehicle and System Technologies'' as part of the AVEAS research project (\href{https://aveas.org/}{\color{black}{www.aveas.org}}).}
}

\author{%
\IEEEauthorblockN{
Florian L\"uttner\IEEEauthorrefmark{2}, %
Nicole Neis\IEEEauthorrefmark{1}, %
Daniel Stadler\IEEEauthorrefmark{3}, %
Robin Moss\IEEEauthorrefmark{2}, %
Mirjam Fehling-Kaschek\IEEEauthorrefmark{2},\\%
Matthias Pfriem\IEEEauthorrefmark{4}, %
Alexander Stolz\IEEEauthorrefmark{5}, %
and Jens Ziehn\IEEEauthorrefmark{3} %
}
\IEEEauthorblockA{\IEEEauthorrefmark{2}Fraunhofer EMI, 
Email: \{florian.luettner, robin.moss, mirjam.fehling-kaschek\}@emi.fraunhofer.de}%
\IEEEauthorblockA{\IEEEauthorrefmark{3}Fraunhofer IOSB, 
Email: \{daniel.stadler, jens.ziehn\}@iosb.fraunhofer.de}%
\IEEEauthorblockA{\IEEEauthorrefmark{1}Porsche Engineering Group GmbH, 
Email: nicole.neis@porsche-engineering.de}%
\IEEEauthorblockA{\IEEEauthorrefmark{4}PTV Group, 
Email: matthias.pfriem@ptvgroup.com}%
\IEEEauthorblockA{\IEEEauthorrefmark{5}INATECH, Albert-Ludwigs-University Freiburg, 
Email: alexander.stolz@inatech.uni-freiburg.de}%
}

\maketitle

\begin{abstract}

The reliable operation of autonomous vehicles, \acl{ad} functions, and \acl{adas} across a wide range of relevant scenarios is critical for their development and deployment. 
Identifying a near-complete set of relevant driving scenarios for such functionalities is challenging due to numerous degrees of freedom involved, each affecting the outcomes of the driving scenario differently. Moreover, with increasing technical complexity of new functionalities, the number of potentially relevant, particularly ``unknown unsafe'' scenarios is increasing. To enhance validation efficiency, it is essential to identify relevant scenarios in advance, starting with simpler domains like highways before moving to more complex environments such as urban traffic.
To address this, this paper focuses on analyzing lane change scenarios in highway traffic, which involve multiple degrees of freedom and present numerous safety-relevant scenarios. We describe the process of data acquisition and processing of real-world data from public highway traffic, followed by the application of criticality measures on trajectory data to evaluate scenarios, as conducted within the AVEAS project (\href{https://aveas.org/}{\color{black}{www.aveas.org}}). By linking the calculated measures to specific lane change driving scenarios and the conditions under which the data was collected, we facilitate the identification of safety-relevant driving scenarios for various applications.
Further, to tackle the extensive range of ``unknown unsafe'' scenarios, we propose a way to generate relevant scenarios by creating synthetic scenarios based on recorded ones. Consequently, we demonstrate and evaluate a processing chain that enables the identification of safety-relevant scenarios, the development of data-driven methods for extracting these scenarios, and the generation of synthetic critical scenarios via sampling on highways.

\end{abstract}

\section{Introduction}

The automation of road traffic is advancing rapidly. While a plurality of trial and test operations can be observed for urban areas worldwide, the introduction of more advanced \acf{ad} functions in mass production vehicles remains mostly focused on highway operation~\cite{ADAC}. This progression is not surprising, considering the paramount importance of the validation of functional safety prior to the homologation and market introduction of new \ac{ad} functions. Highway traffic, in this context, is the least complex and thus the most manageable domain.
A key aspect of the validation challenge is the fact that the real-world driving task is mostly dominated by standard tasks like lane keeping and management of the distance to a leading vehicle, which are not too challenging for the automated system. The true challenge arises from the variety of potential situations that need to be covered safely by the driving logic, including some that occur only very rarely. Statistical reasoning has shown that it would take more than eleven billion miles of driving to prove with high statistical confidence that the failure rate of an automated vehicle is lower than the human driver failure rate~\cite{RAND}. Hence, methods are needed to significantly reduce test mileage and increase the amount of new information per test mile. A common approach is to exclude phases with little test relevance as well as repetitions, and focus on selected distinct scenarios. To speed up development and validation of safe driving functions, it is vital to identify as many relevant scenarios as possible for a broad coverage to ensure safe operation in those. An important measure for the relevance of a potential test scenario is the criticality of the driving situation, where the criticality assessment itself can be based on the combination of several metrics.
To build up a powerful test scenario database, sufficient data from real traffic and thorough analyses are required to pinpoint safety-critical scenarios relevant for the safety of \acf{adas} and \ac{ad} systems. 
To maximize the time efficiency of expanding this data set, derived from empirical observation and analysis, it can be enhanced through both additional data collection and the synthetic generation of further critical maneuvers.

\subsection{Data acquisition and processing from traffic observation} 

The acquisition of real-world data for extracting critical scenarios has gained additional attention with the growing need to validate \ac{ad} functions. Pioneering driving datasets include NGSIM \cite{coifman2017critical_ngsim} and SHRP2 \cite{hankey2016description} consisting of external camera data and in-car data, respectively. 
As presented in \cite{haselbergerJUPITERROSBased2022}, a wide range of vehicle platforms has been developed for the purpose of recording traffic situations from an in-vehicle perspective, differing in the composition and mounting of the sensor suite, and used to record, for example, the well-known datasets KITTI \cite{Geiger2013IJRR} and A2D2 \cite{geyer2020a2d2}. 
The pivotal project PEGASUS \cite{winner2019pegasus} introduced key concepts for the quantitative and scenario-based validation of \ac{ad} functions \cite{weber2019framework}. Subsequent projects have continued along this direction to establish principles on evaluating required data characteristics and acquisition goals \cite{fischer2023pegasus, bein2023verification}. In recent years, efforts have shifted towards assuring the systematic acquisition of such data with the dedicated goal to acquire safety-critical, naturalistic driving behavior and scenarios. Two examples are the KIsSME project \cite{haring2021framework}, which focuses on the development of methods to acquire data through fleets of automated vehicles, and the AVEAS project \cite{eisemann2023approach, eisemann2024joint}, which aims to harmonize efforts for dedicated acquisitions across road vehicle-based, infrastructure-based, and aerial data, mapping each to a common set of file formats (using the ASAM OpenDRIVE and OpenLABEL standards, connected by the joint PAS/standard DIN SAE SPEC 91518).

\subsection{What is ``safety-critical'' road traffic?}

When identifying safety-critical scenarios in road traffic, the term ``critical'' is described and used in various ways in the literature and should be distinguished from the term ``risk'', which is also defined in multiple ways.
In addition to well-established standards for defining risk in the development of \ac{adas} and \ac{ad} functions, such as ISO 26262~\cite{iso26262} and the ISO 31000~\cite{iso31000} Level A standard, several approaches exist for estimating accident risk for individual road users.

According to Junietz et al.~\cite{junietz}, assessing accident risk requires not only information about the driving dynamics of the vehicles involved and current motion parameters such as position, speed, and acceleration, but also insights into the driver's skills and condition, as well as environmental influences. Such data is often scarce in existing datasets and challenging to collect consistently and unambiguously. 
Junietz et al.\ therefore distinguish between accident risk and safety-criticality based on the data used for assessment. 

Graab et al.~\cite{graab} describe critical traffic scenarios as resulting from failures at various cognitive levels of interacting drivers. 
These levels define the process from having access to an information, acquire an information, process the information, derive decisions from the processed information, and act according to the decision made.
This work adopts these definitions.
The analysis focuses on data recorded via vehicles' on-board sensors or from bird's eye view, without considering drivers' abilities or data from external sensors. Moreover, in the recorded data, critical scenarios can only be identified at the level where the observed traffic participants act.
Note that these approaches do not establish a complete independence of accident risk and criticality. With increasing criticality, the risk for an accident increases, and vice versa. 

A more detailed description of approaches for criticality assessment of road traffic can be found in Lüttner et al.~\cite{luttner2025}.

\subsection{Identifying relevant safety-critical maneuvers on highways}

Given the strong correlation between accident risk, frequency, and criticality, insights for identifying relevant maneuvers in critical scenarios can be derived from accident data~\cite{bundesamt2023verkehrsunfalle}. According to Tab.~46241-01 in this statistic, most accidents on highways occur during the car following maneuvers, summing up to $51.46\,\%$ of all accidents.
In addition to following maneuvers, lane changing is one of the most frequently executed maneuvers on highways. 
This maneuver represents the second highest percentage of accidents in Germany, accounting for $44.34\,\%$ of incidents that involve vehicles leaving the road to the right or left. When excluding this road-leaving maneuver, lane changes still contribute to $17.26\,\%$ of accidents. Furthermore, lane changes are particularly more dynamic and complex compared to following traffic due to the various degrees of freedom involved in their execution, which increases the likelihood of safety-critical scenarios in this context.
Consequently, this work focuses on the analysis of lane-changing maneuvers on highways and examines which characteristics of these maneuvers influence their criticality.

For assessing criticality based on the recorded and utilized data, deterministic criticality metrics that capture externally observable driving dynamics (action level, according to Graab et al.~\cite{graab}) are appropriate evaluation measures. A collection of such metrics has been described by Westhofen et al.~\cite{westhofen}. These metrics are widely used for criticality assessment in various studies within the literature, including works~\cite{westhofen, neurohr, lin, kondoh, junietz, junietz2}. Additionally, in a study by Lüttner et al.~\cite{luttner2025}, a comprehensive evaluation framework is proposed as a new approach, which uses the metrics collection by Westhofen et al. This study also suggests threshold values for these metrics to aid in identifying critical road traffic scenarios on highways.

The following chapter outlines the process of data acquisition from two distinct recording perspectives (\cref{sec:data}). Following that, the identification of lane changes within the recorded data and the examination of the robustness of the identification methods is described (\cref{sec:lc}). Additionally, the analysis of relevant metrics, referenced by their criticality, is presented, along with their interpretation in relation to identifying critical lane change maneuvers derived from self-recorded and analyzed data and criticality-referenced sampling of lane changes in the traffic flow simulation environment PTV Vissim (\cref{sec:crit}). Finally, conclusions are drawn (\cref{sec:conclusion}).

\section{Applied Data Acquisition Methods} 
\label{sec:data}

Lane change data at highways was acquired from two sources: direct local measurements by a research road vehicle (\cref{sec:in-car}), and aerial footage from an ultralight aircraft equipped with camera and position sensors (\cref{sec:aerial}).

\subsection{In-car data acquisition}\label{sec:in-car}

In-car data is collected from the Porsche Engineering (PEG) research vehicle platform JUPITER~\cite{haselbergerJUPITERROSBased2022}, equipped with multiple sensor systems in addition to the series vehicle equipment, including LiDAR systems to the front, back and the sides, a stereo camera, and GNSS with RTK for accurate localization of the vehicle in road traffic. 
Compared to other state-of-the-art recording platforms, the one of the JUPITER platform is closest to series production vehicles \cite{haselbergerJUPITERROSBased2022}. Due to the inclusion of data provided by the series bus such as the distances to the left and right lane markings into the recordings, it further allows accurate in-lane positioning that lacks, for example, in the KITTI and A2D2 datasets. The aforementioned signals that trace back to the series front monocular camera system are leveraged for the recognition of lane changes. 
Speed information used in subsequent steps is obtained from the GNSS inertial system.  
The signals are provided at frequencies of \num{25}\,\si{\hertz} (bus) and \num{100}\,\si{\hertz} (GNSS inertial system) but are down-sampled and synchronized to  \num{5}\,\si{\hertz}. Further, frequencies above \num{1.3}\,\si{\hertz} are removed with a low-pass filter, as from this frequency onward, measurement inaccuracies, e.g., from the periodicity of the dashed lane marking, determine the behavior rather than human driving behavior. 

Lane change data were collected on three different drivers ($D_1$, $D_2$, $D_3$) on German highways near Stuttgart, capturing the influence of different driving behaviors and situations on the lane change characteristics. The recording drives had a total duration of \num{9.4}\,\si{\hour} for \emph{$D_1$}, \num{7.5}\,\si{\hour} for \emph{$D_2$}, and \num{6.1}\,\si{\hour} for \emph{$D_3$}.

\subsection{Aerial data acquisition}\label{sec:aerial}
Aerial image recording is based on the extraction of geo-localized trajectories with precise lane positions of vehicles.
Data was acquired with an ultralight aircraft above the German highway A5 near Karlsruhe on seven different dates.

The sensor platform includes a calibrated camera capturing images at \num{20}\,\si{\hertz} with a resolution of $4104\times3006$\,px and a ground sampling distance of approximately 12.5\,cm, an \ac{imu} and a GNSS.
To detect traffic participants, i.e., \emph{cars} and \emph{trucks}, a ReDet~\cite{redet} model with ReResNet50 backbone, initialized with DOTA~\cite{dota} pre-trained weights, was finetuned on 1\,324 manually annotated images from the target domain, using the MMRotate~\cite{mmrotate} library.
A \ac{map} of 87.9 on 510 test images indicates a high detection accuracy of the applied model.
Note that ReDet outputs oriented bounding boxes such that the length and width of a detected vehicle can be directly derived, in contrast to standard detectors that yield axis-aligned boxes.

For analyzing lane changes, accurate positions of the observed vehicles w.r.t.\ lane markings are required, which are obtained as follows.
First, a \ac{dem} is computed using the \ac{imu} and GNSS data next to the images via bundle adjustment.
Then, an orthophoto of the area is created from the DEM.
This allows easy manual annotation of lanes and consistent geo-referencing of detected vehicles.

\begin{figure}
    \includegraphics[width=\columnwidth]{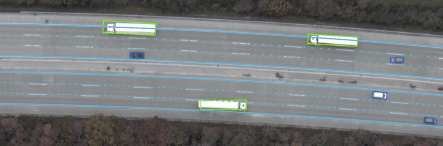}
    \vspace{-3mm}
    \caption{Aerial image patch with detected cars, trucks, and annotated lanes. It can be observed that in spite of the highly accurate geo-referencing, the bounding box positions of large vehicles exhibit a lateral bias in the lane based on perspective effects.}
    \label{fig:aerial_data_acquisition}
\end{figure}

\begin{figure*}
    \begin{subfigure}{0.32\linewidth}
    \includegraphics[width=\columnwidth]{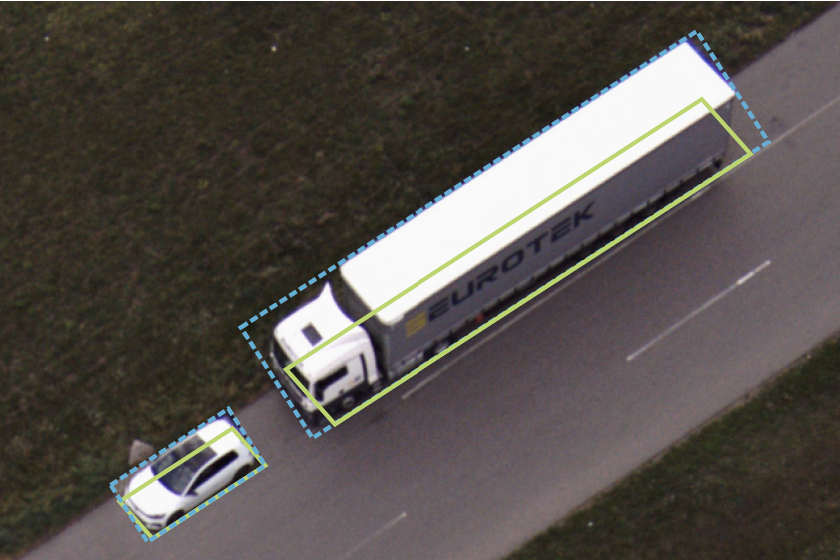}
    \caption{extreme perspective with bounding boxes}
    \end{subfigure}
    \hfill
    \begin{subfigure}{0.32\linewidth}
    \includegraphics[width=\columnwidth]{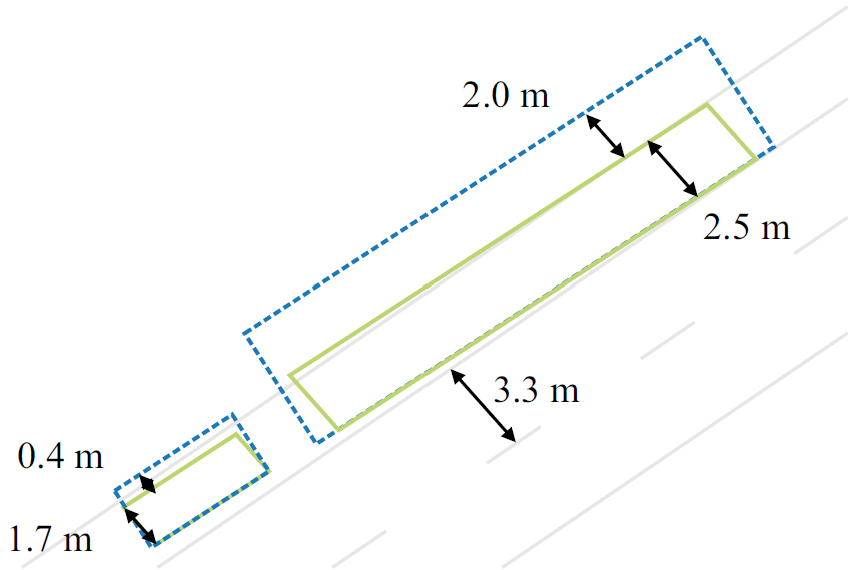}
    \caption{bounding boxes (dashed) vs. footprints (solid)}
    \end{subfigure}
    \hfill
    \begin{subfigure}{0.32\linewidth}
    \includegraphics[width=\columnwidth]{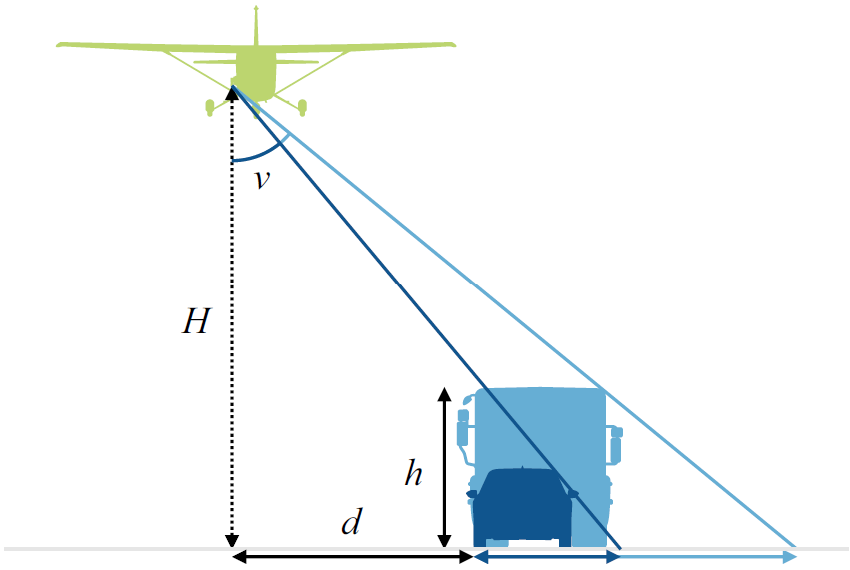}
    \caption{perspective effect contributions}
    \end{subfigure}\\

    \caption{Object detection in aerial images through oriented bounding boxes can lead to unilateral overestimation of object size in range direction through perspective effects, and thus imply erroneous overlaps with other lanes (shown here as an extreme example at steep off-nadir angles $\nu$ and low altitudes $H$ during final approach). Effects depend on off-nadir distances, object heights $h$, but also object vertical shape. While bounding box precision is high ($\approx$ 0.2\,m), the constant per-object bias can be considerably larger ($\approx$ 1\,m under described regular acquisition conditions at cruising altitudes, cf. \cite{eisemann2023approach}).}
    \label{fig:aerial_perspective}
\end{figure*}

Tracking of the traffic participants is done in orthophoto coordinates, which renders an additional camera motion compensation component in the tracker obsolete.
Due to the high frame rate of the camera, a simple motion-based tracker using a Kalman filter implementation from \cite{deepsort} and \acl{iou} of oriented bounding boxes as association similarity is applied.
The resulting trajectories are geo-localized and contain precise lane positions at each timestep. However, for low altitudes, perspective effects of dynamic objects that are robustly extracted in the camera image are not generally negligible (cf.\ \cref{fig:aerial_data_acquisition,fig:aerial_perspective}), nor can they be compensated in the bundle adjustments (in contrast to perspective effects of the static terrain). For this reason, trajectories are highly precise (with high spatial resolution and low noise), but prone to exhibiting a low-frequency per-vehicle bias. The generated aerial trajectory dataset comprises 53\,476 different vehicles from a total recording time of 5 hours and 9 minutes.

\section{Robust Lane Change Statistics}
\label{sec:lc}

To compare lane change durations between different acquisition methods, a metric is required that is sufficiently interpretable, while yielding robust results in both acquisition modes (in-car and aerial).
Commonly used for identifying the duration of a lane changes is the midpoint of a lane change, which is set to the point in time when the lane marking is crossed.
Literature is rather unanimous regarding the definition of the midpoint of a lane change \cite{biparvaVideoActionRecognition2022, benterkiMultiModelLearningBasedFramework2020}. However, opinions differ on the determination of the start and end of a lane change. Proposed approaches include the definition of a fixed time window \cite{benterkiMultiModelLearningBasedFramework2020} or spatial area \cite{assadiFlexibleStochasticMicroscopic2020} around the lane change's midpoint, and the definition via the start and end of lateral movement \cite{butakovPersonalizedDriverVehicle2015}; according to the UN regulation 79, a lane change maneuver starts with one wheel overlapping with the lane marking and ends when both wheels have crossed it\cite{AgreementConcerningAdoption2018}.

These definitions require, when applied analytically, specific information that is not generally available in measurement data with sufficient accuracy. Wheel positions are not extracted from aerial images, and lateral motions of vehicles within their lane (even under highly accurate recording conditions) are too significant also during lane following to sharply define maneuver start and end points under realistic highway conditions. Additionally, the perspective effects in the aerial data (\cref{sec:aerial}) imply the need for a metric that is robust against a low-frequency lateral bias (as introduced by the perspective effects in the aerial data). However, beside the perspective effects both data sources provide the in-lane position with relatively high precision; thus, robustness against excessive noise is not required.
For the in-car recordings, lane changes can be identified using the method presented in \cite{montanariPatternRecognitionDriving2020}, which detects lane changes based on large gradients in the distances to the lane markings that arise from the change of the referenced left respectively right lane marking when crossing a lane marking. While this approach is highly reliable, its applicability is limited to acquisition methods providing the lane marking distance information in the required form, and therefore in particular not suitable for aerial recordings. 

\begin{figure}[t]
    \centering\includegraphics[width=0.4\textwidth]{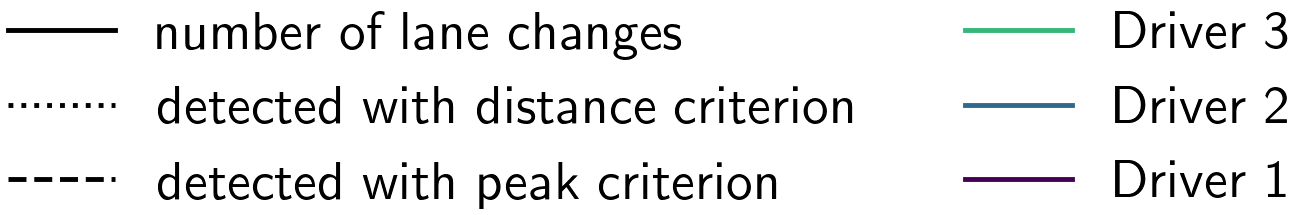}
    \includegraphics[width=0.3\textwidth]{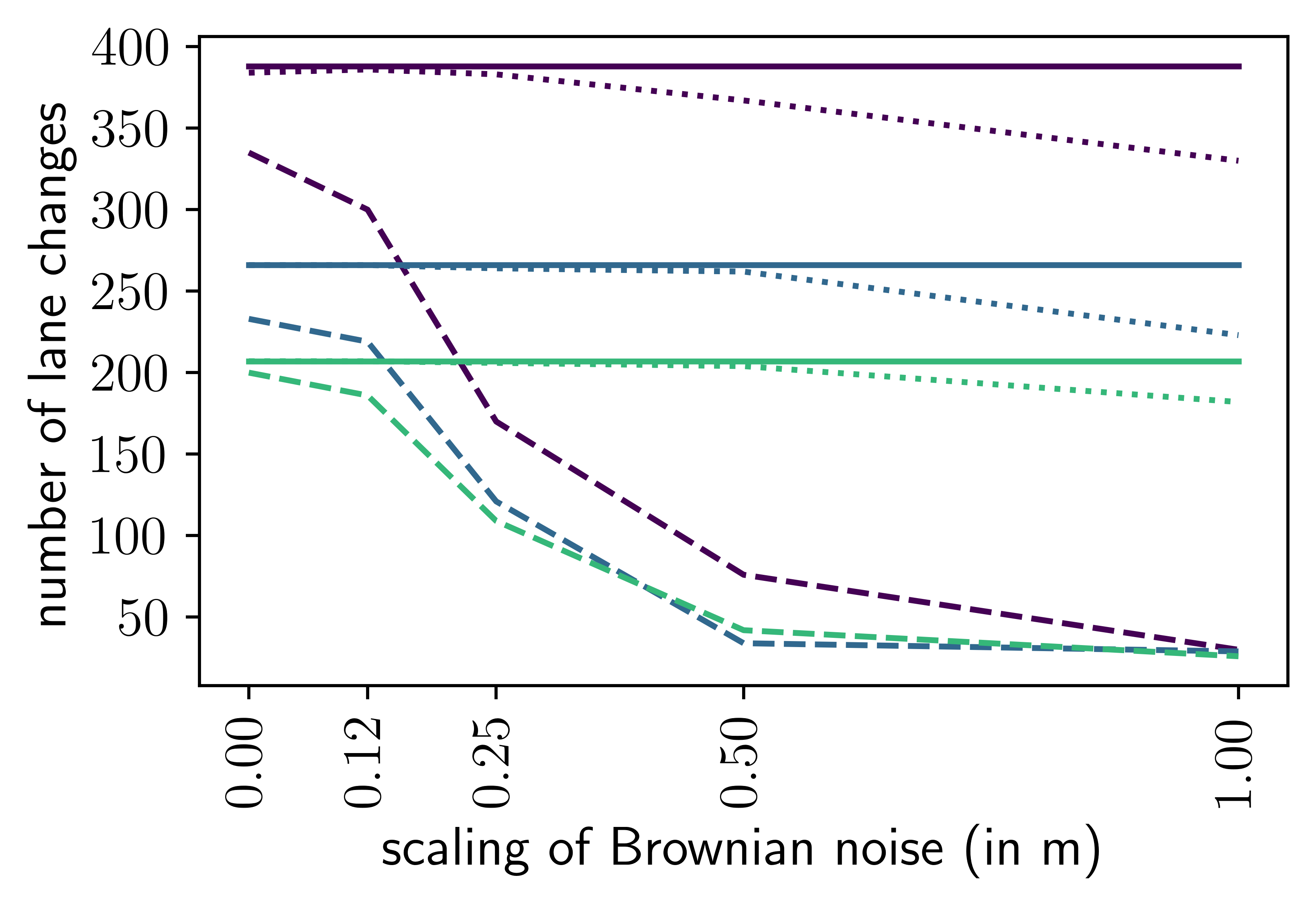}
    \includegraphics[width=0.3\textwidth]{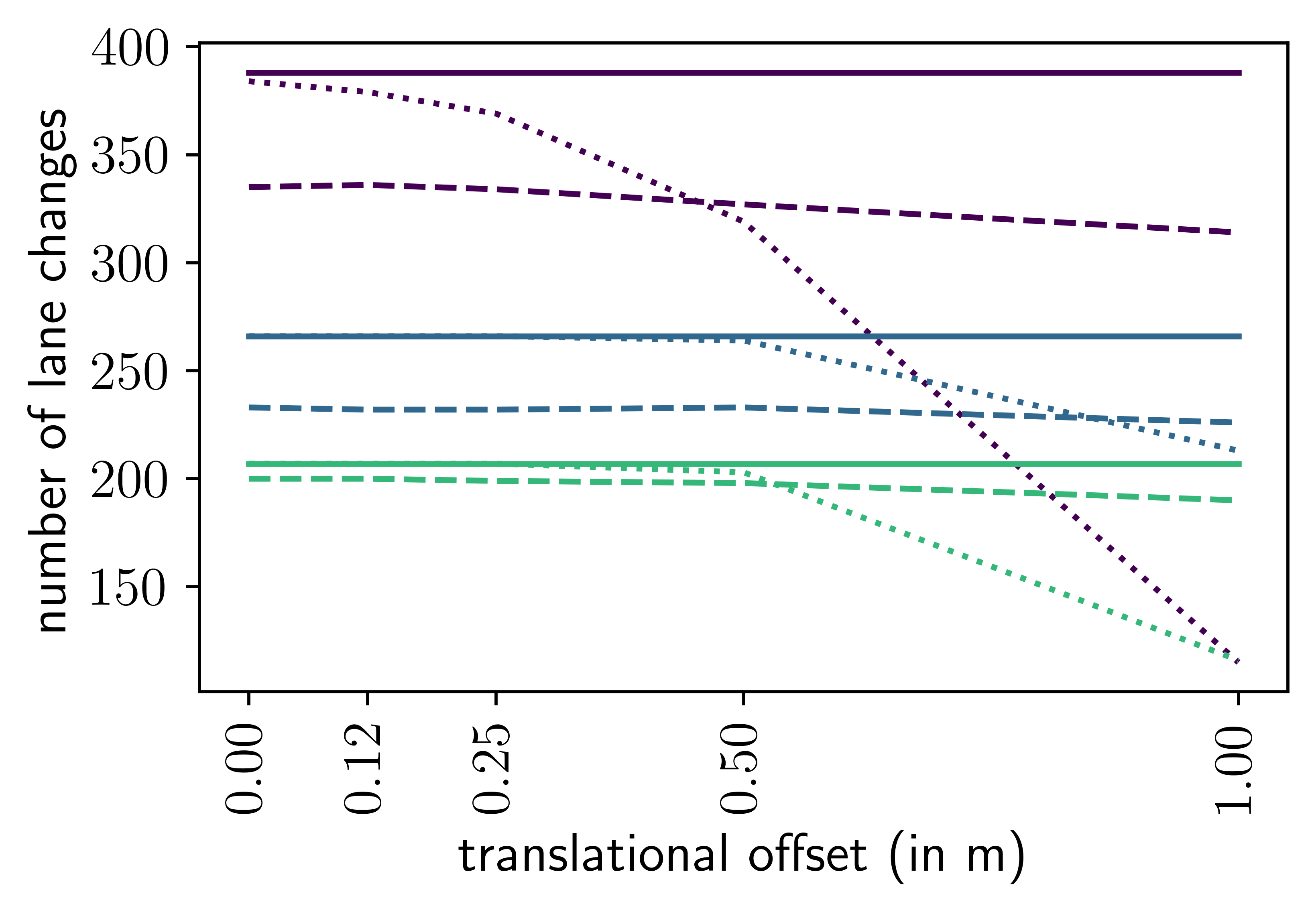}
\caption{\label{fig:dist_peak_peg}
 Comparison of the number of detected lane changes using the distance criterion (dotted lines) and the peak criterion (dashed lines). For comparison, the actual number of executed lane changes (ground truth in solid lines) is presented. The comparison was conducted for three drivers ($D_1$ in purple, $D_2$ in blue, $D_3$ in green) of the JUPITER vehicle. The number of detected lane changes was examined depending on artificially imposed inaccuracies on the measurement results in the form of Brownian noise (top) and translations (bottom) regarding the vehicle's position determination.}
\end{figure}

In the following, two metrics for lane change identification are compared. These are referred to as \emph{distance criterion} and \emph{peak criterion}. The approach based on distance gradients serves as ground truth for reference.
In case of the distance criterion, a lane change is detected through a sufficiently large lateral displacement from the lane center, similar to \cite{assadiFlexibleStochasticMicroscopic2020}.
The peak criterion detects the lane change through peaks in the derivative of the lateral position, using the \texttt{find\_peaks} function as part of the scipy signal package \cite{2020SciPy}. The subsequent duration is then calculated based on the width of the detected peak using the \texttt{peak\_widths} function of the same package \cite{2020SciPy}, where the relative percentage height at which the lane change duration is measured is calculated as $\text{rel\_height} = 1 - (\text{width}\subs{obj}/\text{width}\subs{lane})/2$. 
Both criteria were evaluated for robustness against Brownian noise and a constant bias by applying these uncertainties to in-car recordings, where the total number of lane changes was accurately known from gradient identification. The number of identified lane changes was then compared to the known ground truth for each method. 
The results shown in \cref{fig:dist_peak_peg} indicate that the peak criterion is strongly affected by noise, but far more robust against a constant translational offset. 
Additionally, the peak criterion proves to be more accurate in detecting lane changes for larger translational offsets than the distance criterion. These values are particularly relevant when considering trucks from an aerial perspective, where the oblique view and height of the trucks can affect detection accuracy.

\subsubsection*{Peak lane change detection and measurement}\label{sec:peaks}

As aerial data feature errors from perspective bias (\cref{sec:aerial,fig:aerial_perspective}), from the previously performed experiments, the peak criterion is considered more suitable and applied in the following to detect lane changes: Systematic long-term errors, especially from perspective overlaps (in the order of 1\,m and above) exceeds notably the short-term inaccuracies (in the order of 0.2\,m), such that the peak criterion yields more robust results. As part of the following analysis, only fully recorded lane changes, which had a width of at least \num{2.5}\,m, were considered in the analysis of the aerial data.

\begin{figure*}
\begin{subfigure}[c]{0.95\textwidth}
        \includegraphics[width=\textwidth]{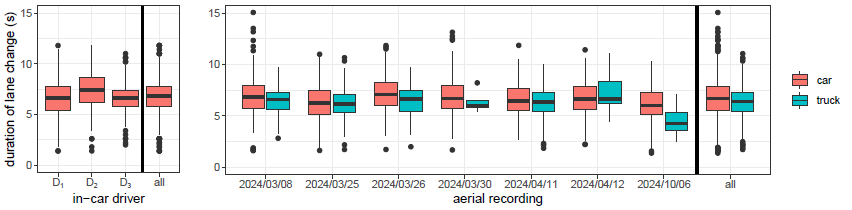}
    \label{fig:dist2_dur}
\end{subfigure}\\[-5mm]
\begin{subfigure}[c]{\textwidth}
    \includegraphics[width=0.95\textwidth]{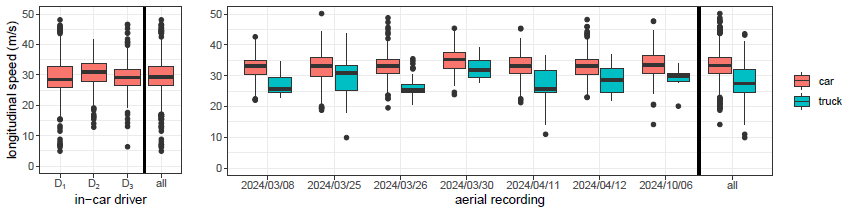}
    \label{fig:dist2_speed}
\end{subfigure}
\caption{\label{fig:dist2_all} Lane change durations (top) and speed at the peak point of the lane crossing (bottom) for lane changes of cars (red) and trucks (blue) observed in in-vehicle (left) and bird's eye view (right) data. Median values of all identified lane changes for each driver or recording (horizontal lines) are depicted within the left side of each figure, and the summary of all recordings on the rightmost entry (separated by a vertical solid line) with 25 and 75 percentiles (blue and red boxes), maximum range (whiskers), and outliers (black circles). Dates for the aerial recording are given on the x-axis in the year/month/day format.}
\end{figure*}

In total, 857 lane changes were recorded from the in-car drivers and 1\,615 lane changes were identified in the aerial recordings. In addition, 30 double lane changes (from the outer left or right lane to the other outer right or left lane on a three lane highway) were identified in the aerial recordings and 39 for the in-car drivers. Due to the low number of double lane changes, they are not considered in the following. The estimated durations and speed values at the peak of the crossing are shown in \cref{fig:dist2_all}. In summary, the results obtained by both recording methods (in-car and aerial) are in good agreement: mean durations of lane changes of 6.87\,s (in-car) and 6.73\,s (aerial) and mean speed values of 32.5\,m/s (in-car) and 33.2\,m/s (aerial) were observed for cars.
As expected, slightly lower speed values were observed for trucks (mean speed: 28.3\,m/s) compared to cars. Interestingly, the mean lane change duration of trucks (6.36\,s) is also observed slightly below the duration for cars. 
The observed lane change durations and speed values were found to be statistically uncorrelated.
In general, both types of recordings are suitable for lane change analysis. Although the in-car recorded data closely match in terms of speed and lane change duration, the aerial data display a higher statistical variance in the composition of vehicle classes and driver types. The in-car data were restricted to only three drivers, all operating the same vehicle (JUPITER by PEG). To eliminate potential bias in the evaluation, the following metric considerations were exclusively applied to the aerial data.

\section{Criticality Analysis of Lane Changes}
\label{sec:crit}

A special interest in the analysis of lane changes is determining if the lane change leads to safety-critical situations. 
The criticality of the lane changes was assessed by applying a set of metrics as described in~\cite{luttner2025}. 
A summary of the selected metrics is provided in \cref{tab:metrics}.

\begin{table}
\vspace{0.5cm}
\caption{\label{tab:metrics} Summary of criticality metrics based on~\cite{luttner2025}, providing their symbol, name, and threshold value above or below which the value is considered critical. The parameter $\vlim$ in the threshold definition of ${v}$ corresponds to the speed limit of the selected street segment.
}

\begin{center}
\renewcommand{\arraystretch}{1.2}\small
\begin{tabular}{|c|c|c|c|c|c|c|}
\hline
\textbf{Symbol} & \textbf{Name}   & \textbf{Threshold} \\
\hline
 $d$ & Euclidean distance   & $d < 1.0\,\si{\metre}$ \\
\hline
 $v$ & longitudinal velocity   & $v > 1.3 \cdot \vlim $ \\
\hline
 $a\subs{lon}$ & longitudinal acceleration  & $a\subs{lon} > 8\,\si{\metre/\second^2}$ \\
\hline
 $a\subs{lat}$ & lateral acceleration   &  $a\subs{lat} > 8\,\si{\metre/\second^2}$ \\
\hline
 $\THW$ & time headway   & $\THW < 0.9\,\si{\second}$ \\
\hline
 $\DCE$ & distance of closest encounter   & $\DCE < 1.0\,\si{\metre}$ \\
 \hline
 $\TTCE$ & time to closest encounter   & $\TTCE < 2.6\,\si{\second}$\\
\hline

\hline
\end{tabular}
\end{center}
\vspace{-0.5cm}
\end{table}
The threshold of each metric defines the range of values considered critical: below the threshold for $d$, $\THW$, $\DCE$, and $\TTCE$, and above the threshold for $v$, $a\subs{lon}$, and $a\subs{lat}$. 
The metrics $\DCE$ and $\TTCE$ cannot be considered standalone, as a small $\DCE$ is only relevant if the closest distance will be reached in a very short time, i.e., a small $\TTCE$, and vice versa. Therefore, the $\DCE$ values were evaluated only for data with $\TTCE < 2.6\,\si{\second}$. 
The threshold of the velocity metric requires the value of the speed limit of the recorded roads. Only parts of the highway have speed limits, and as a conservative estimate, we set $\vlim=120$\,km/h for the full dataset.

\begin{figure*}
    \centering
    \includegraphics[width=1\textwidth]{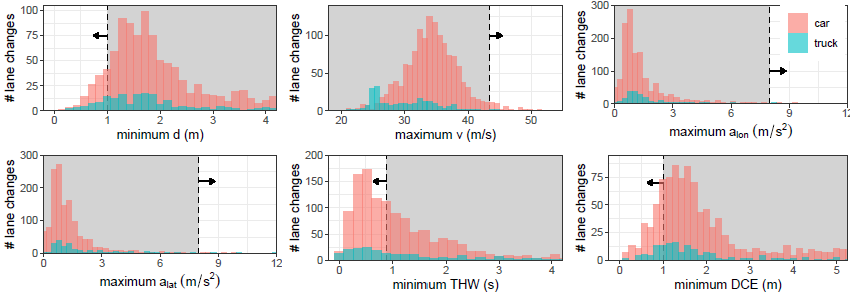}
    \vspace{-6mm}
    \caption{Most critical metric values of lane changes based on the evaluation of criticality metrics for the aerial recordings. Minimum metric values were calculated for $d$, $\THW$, and $\DCE$ and maximum metric values for $v$, $a\subs{lon}$, and $a\subs{lat}$. $\DCE$ values were only computed when $\TTCE < 2.6$\,s. Criticality thresholds are shown by dashed vertical lines. Non-critical areas are shaded in gray. Black arrows indicate the direction towards higher critical values.}
    \label{fig:hist_metrics}
\end{figure*}

All metrics are evaluated from the ego perspective of the lane-changing vehicle, considering all other recorded vehicles in the same time frame as opponents.
In the case of critical metric values from several opponent vehicles, the most critical value was selected. 
The distributions of the most critical metric values for the lane-changing vehicles are shown in \cref{fig:hist_metrics} for the aerial data. 
As can be seen, only a small number of lane changes are detected in the critical regions for most metrics. However, for $\THW$, a significant fraction of lane changes is found in the critical region. 

\begin{figure*}
    \centering
    \includegraphics[width=1\textwidth]{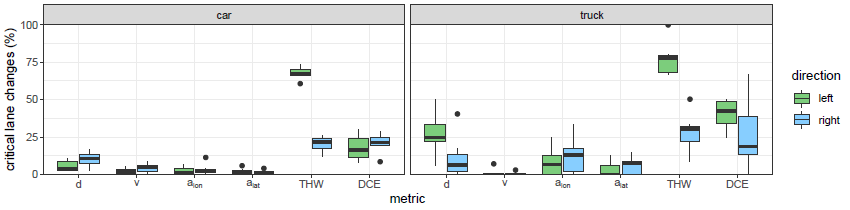}
    \vspace{-6mm}
    \caption{Percentage of critical lane changes based on the evaluation of criticality metrics for the aerial recordings. The percentage of critical lane changes was computed per recording day and direction (green: left, blue: right). The median values of all recordings (horizontal lines) are displayed with 25 and 75 percentiles (boxes), maximum range (whiskers), and outliers (black circles).}
    \label{fig:percent_crit}
\end{figure*}

Furthermore, it was investigated whether lane change parameters such as duration or speed are correlated to any of the metrics. The speed is directly correlated to the $v$ metric and other metrics containing velocity information such as $\THW$. Note that no correlation to the metrics was found for the lane change duration but a significant difference was observed for $\THW$ values for the lane change direction (left or right). 
This is shown in \cref{fig:percent_crit}, where the percentage of critical lane changes is displayed for all metrics, separately for left and right lane changes.
While critical $\THW$ values are observed for 68.5\,\% of left lane changes, only 22.2\,\% of right lane changes have critical $\THW$ values.
The difference can be explained by the cut-out and cut-in maneuvers during overtaking. Initially, a left lane change (cut-out) occurs, followed by a right lane change (cut-in). During the cut-out, the vehicle accelerates to overtake the slower vehicle and potentially close up to a proceeding vehicle after the cut-out, resulting in critical $\THW$ values due to a shorter following distance and higher speed, despite the distance and speed being non-critical at that moment.
Only small differences within statistical significance are observed between left and right lane changes for the other metrics, which generally show lower percentages of critical lane changes.
Note that the chosen metrics are suitable for identifying both critical lateral maneuvers and maneuvers in straight-line traffic. Specifically, $\THW$, the distance $d$, the combination of $\TTCE$ and $\DCE$ and longitudinal acceleration $a_{lon}$ are suitable for detecting critical, non-lane-based maneuvers.


\subsection{Sampling}
\begin{figure*}
    \centering
    \includegraphics[width=0.7\textwidth]{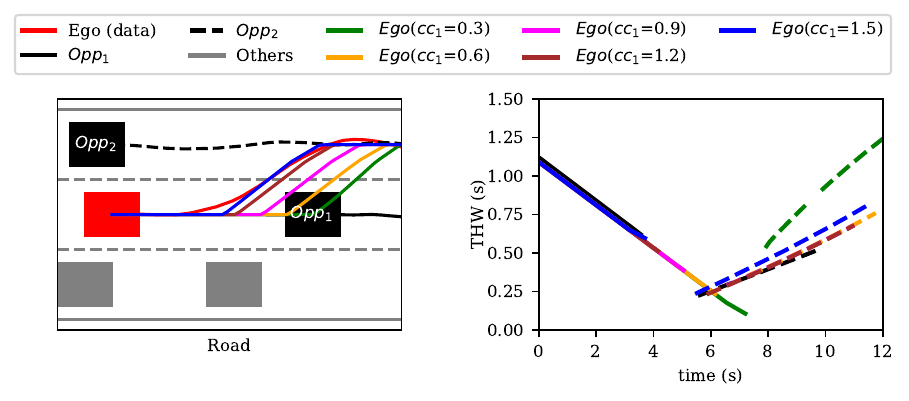}
    \vspace{-3mm}
    \caption{Left: Exemplarily sampled scenario of an ego vehicle (red) being overtaken by one opponent ($\textit{Opp}_2$) and overtaking another opponent ($\textit{Opp}_1$) afterwards. Right: Respective $\THW$ metric of the ego vehicle vs.\ different opponents (solid: $\textit{Opp}_1$, dashed: $\textit{Opp}_2$) for different $cc_1$ values. (Vehicle lengths not to scale.)}
    \label{fig:sampling}
\end{figure*}

As part of the processing chain, the detected lane change scenarios can then be sampled to create new partially synthetic scenarios.
In order to demonstrate such sampling, we manually selected one of the detected lane change scenarios. 
Subsequently, the trajectories of the respective vehicles of said scenario were imported into PTV Vissim \cite{VissimRef} with one vehicle being replaced by a Wiedemann99 behavior model.
The $cc_1$ parameter of the Vissim-controlled vehicle, which represents the target gap in seconds to the preceding vehicle, was then varied to generate new scenarios.
The result of such variation and difference in the respective calculated criticality can be seen in \cref{fig:sampling}.
In the original scenario, the $\textit{Ego}$ (red) vehicle is overtaken by $\textit{Opp}_2$ (black, dashed) and overtakes the $\textit{Opp}_1$ (black, solid) vehicle afterward.

In general, infinite sampled scenarios with varied parameter values can be generated for each observed critical scenario. However, to observe a significant difference in the resulting simulated scenario, a discretization of the steps-size of the parameter variation to 5 to 10 variations is applied. Therefore, different trajectories are depicted in \ref{fig:sampling}, where the parameter $cc_1$ is sampled in 5 steps towards lower values, causing a reduction in the gap and consequently $\THW$ between $\textit{Ego}$ and $\textit{Opp}_1$.
In contrast, the $\THW$ between $\textit{Ego}$ and $\textit{Opp}_2$ is largely unaffected, as $\textit{Opp}_2$ has a higher speed than the $\textit{Ego}$ vehicle.

\begin{figure*}
    \centering
    \includegraphics[width=\textwidth]{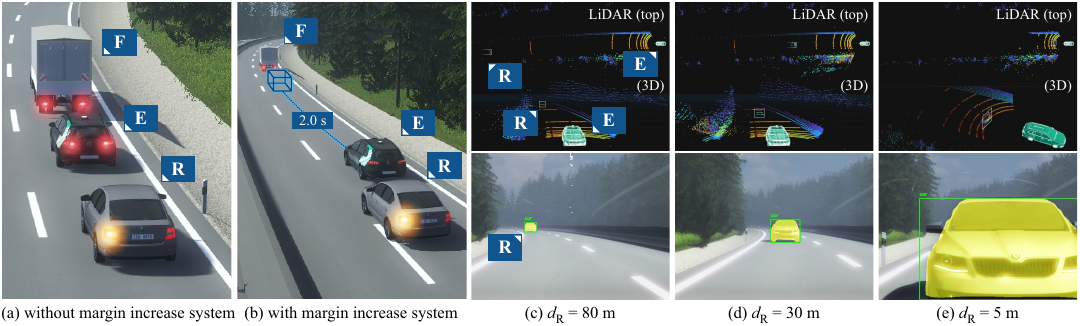}
    \caption{Application of evaluating the MIS (\cref{sec:example}) from sampled lane changes. The system anticipates overtaking maneuvers with low rear THW (a), increasing the front THW to mitigate braking risks. (b--e) show the sampled simulation scenario for a rear THW of 0.2\,s. The automated ego vehicle \textbf{E} detects the approaching rear vehicle \textbf{R} at a range of 100\,m, decelerates by $1\,\unit{\metre/\second^2}$ to increase its front THW by 2\,s. This way, during the close lane change of \textbf{R} at $d\subs{R} \approx 5\,\unit{\metre}$ (e), \textbf{E} has sufficient margin to not brake until \textbf{R} concludes its maneuver, regardless of any possible braking action of \textbf{F}.}
    \label{fig:example}
\end{figure*}

\subsection{Exemplary Application}\label{sec:example}

The approach is used for simulated testing of a concept \emph{margin increase system} (MIS) for automated driving, implemented in the \octas\footnote{octas.org} simulation framework, based on the described behavior models in a software in the loop setting.  \cref{fig:example} shows an exemplary application. Close overtaking maneuvers (with low $\THW$) pose a considerable risk, as the overtaking vehicle (\textbf{R} / $\textit{Opp}_1$) disregards the risk of the \textit{Ego} vehicle braking unexpectedly. This risk arises during the final stages of the overtaking maneuver, when \textbf{R} is close to \textbf{E}, but not yet fully beside it. If \textbf{E} has to brake, particularly due to its front vehicle \textbf{F} braking, then there is no rear safety margin.

AD vehicles in fully automated mode can select a narrow front margin, based on the comfort limits at very short reaction times of the automated perception system. Such narrow margins (lower than recommended distances for human drivers) improve efficiency and reduce the risk of other traffic participants consistently merging into the forward gap. This operation, however, requires that the AD vehicle can react immediately to actions of the front vehicle. For very close overtaking maneuvers of human drivers, however, a braking of the front vehicle can then cascade to cause a rear-end collision between the AD vehicle \textbf{E} and the rear human driver \textbf{R}. The purpose of the MIS is to mitigate this risk by proactively increasing the front margin beyond the regular cruising limit, if a critical overtaking maneuver is anticipated (cf. \cref{fig:example}).

The shown MIS utilizes a rear LiDAR (Valeo Scala Gen. 2) and a rear monocular camera for vehicle detection and classification, using density-based clustering on the LiDAR point cloud and a Mask R-CNN on the camera data. The MIS engages when the AD \textit{Ego} vehicle \textbf{E} is in fully automated mode and following a front vehicle \textbf{F} at a short margin; when a vehicle \textbf{R} is detected at $d\subs{R} = 100\,\unit{\metre}$ or less rear distance; when \textbf{R} is approaching at a delta speed of 10\,km/h or above; when the left lane is free for the \textbf{R}'s overtaking maneuver. In this case, the $\TTCE$ between \textbf{E} and \textbf{R} is calculated to determine a braking acceleration that increases the front $\THW$ (\textbf{F}$\rightarrow$\textbf{E}) by $+2\,\unit{\second}$. Thereby, \textbf{E} can assure that it does \emph{not} require braking for an interval of two seconds during the most critical part of the overtaking maneuver. The simulation results based on the developed models indicate that the system can reliably increase the margin at convenient decelerations in time (\cref{fig:example}b--e), whereas situations without the MIS (\cref{fig:example}a) can cause critical rear-end collisions through cascaded braking in the sampled low-$\THW$ scenarios.

\section{Conclusions}
\label{sec:conclusion}

The paper illustrates a comprehensive approach to achieving robust data acquisition followed by maneuver extraction and the analysis of critical maneuvers based on relevant metrics. Further analysis on the critical maneuvers is conducted via simulated sampling. The focus is on lane change maneuvers on highways.
The data used for the evaluation and analysis of relevant criticality metrics for lane change behavior on highways, captured from two acquisition methods and perspectives, enables the evaluation of lane change detection methods in practical applications and their transferability.
The comparison of the two applied lane change recognition algorithms clearly shows that the robustness of maneuver identification heavily depends on the selection of the appropriate method relative to the recording type, as well as the inaccuracies arising from data recording and processing. However, by appropriately selecting methods for maneuver extraction while considering identified inaccuracies in data acquisition and processing, similar statistical characteristics for the examined lane changes can be observed from both the ego perspective and the aerial perspective. This indicates a valid extraction of lane change maneuvers as well as the possibility to establish harmonized results from heterogeneous data sources.
Based on this, a comprehensive investigation into lane change-relevant criticality metrics is presented, revealing distinct differences between maneuvers to the right and left.
The analysis identified a higher criticality for left-going lane changes, particularly when examining the $\THW$. 
Simulations have proven to be an effective tool for sampling relevant scenarios that mirror real-world behaviors, addressing the challenge of the scarcity of such scenarios identified in actual data. Furthermore, simulations are used for testing a concept \emph{margin increase system} (MIS) for automated driving, implemented in the \octas{} simulation framework using the generated results.


\subsection*{Outlook}
The insights gained from the performed analysis provide valuable approaches for a deeper understanding of lane change maneuvers on highways. They facilitate the adaptation and optimization of these maneuvers within simulation frameworks, ultimately contributing to the development of safer and more efficient driving systems. This is underscored by the fact that the done investigations align with the current advancements in the automation of road traffic. The study provides indications to take a closer look at direction-dependent lane changes of different road users. Such scenarios and the insights derived from them can be significant for the further development of AD functions that are already in use, such as a highway pilot or a MIS system. This is also shown with simulations for the MIS.
Furthermore, this research underlines the importance of a robust design for \ac{adas} on highways, based on a thorough understanding of lane change investigations and the critical metrics involved. 
Further work must expand on the harmonization of heterogeneous data sources to establish a common body of data that enables calibrating models to local differences in behavior, to improve the understanding of effects in traffic safety, and accelerate reactive measures, and on the quantitative comparison between predicted system limits and their safety impacts, and the true performance of a system under test within its operational design domain.


\bibliographystyle{IEEEtran}
\bibliography{IEEEabrv,bibliography}

\end{document}

%% file: include/includes.tex
\usepackage{amsmath}
\usepackage{amssymb}

\usepackage[table]{xcolor}

\usepackage{tikz}
\usepackage{array}
\usepackage{framed, color}
\usepackage[normalem]{ulem}
\usepackage{cancel}
\usepackage{overpic}
\usepackage[footnotesize]{caption}
\usepackage[font+=footnotesize]{subcaption}
\usepackage[nodisplayskipstretch]{setspace}
\usepackage{txfonts}
\let\mathbb=\varmathbb

\usepackage{url}
\usepackage{upgreek}
\usepackage{float}
\usepackage{booktabs}
\usepackage{courier}
\usepackage[hidelinks]{hyperref}
\usepackage{nicefrac}
\usepackage{xcolor}
\usepackage{mdframed}
\usepackage{cleveref}
\usepackage{acronym}

\usepackage{cite}
\usepackage{amsfonts}
\usepackage{algorithmic}
\usepackage{graphicx}
\usepackage{textcomp}
\usepackage{xcolor}
\usepackage{siunitx}
\usepackage{comment}

\usepackage{multirow}

%% file: include/setup.tex
\renewcommand{\arraystretch}{1.3}

\newcommand{\subs}[1]{_{\textrm{\upshape{#1}}}}

\crefname{figure}{Fig.}{Figs.}
\Crefname{figure}{Fig.}{Figs.}

\crefname{table}{Tab.}{Tabs.}
\Crefname{table}{Tab.}{Tabs.}

\crefname{equation}{Eq.}{Eqs.}
\Crefname{equation}{Eq.}{Eqs.}

\crefname{section}{Sec.}{Secs.}
\Crefname{section}{Sec.}{Secs.}

\crefname{subsection}{Sec.}{Secs.}
\Crefname{subsection}{Sec.}{Secs.}

\newcommand\registered{$^\text{\textregistered}$}
\newcommand\octas{OCTAS\registered}

%% file: include/acronyms.tex
\acrodef{adas}[ADAS]{advanced driver assistance systems}
\acrodef{ad}[AD]{automated driving}
\acrodef{odd}[ODD]{operational design domain}
\acrodef{imu}[IMU]{inertial measurement unit}
\acrodef{dem}[DEM]{digital elevation model}
\acrodef{map}[mAP]{mean Average Precision}
\acrodef{iou}[IOU]{intersection over union}